\documentclass[]{oppo}
\usepackage[toc,page,header]{appendix}
\usepackage{latexsym}
\usepackage[T1]{fontenc}
\usepackage[utf8]{inputenc}
\usepackage{microtype}
\usepackage{inconsolata}
\usepackage{graphicx}
\usepackage{hyperref}       
\usepackage{url}            
\usepackage{booktabs}       
\usepackage{amsfonts}       
\usepackage{nicefrac}       
\usepackage{stackengine}
\usepackage{microtype}      
\usepackage{colortbl}
\usepackage{xcolor}
\usepackage{amsmath}
\usepackage{amssymb}
\usepackage{amsthm}
\usepackage{mathrsfs}
\usepackage{pifont}
\usepackage{MnSymbol}
\usepackage{balance}
\usepackage{enumitem}
\usepackage{listings}
\usepackage{xcolor}
\usepackage{natbib}
\usepackage{multicol}

\AtBeginDocument{%
  \providecommand\BibTeX{{%
    \normalfont B\kern-0.5em{\scshape i\kern-0.25em b}\kern-0.8em\TeX}}}

\makeatletter
\DeclareRobustCommand\onedot{\futurelet\@let@token\@onedot}
\def\@onedot{\ifx\@let@token.\else.\null\fi}

\newcommand{\etc}{\emph{etc\@\onedot}}

\usepackage{setspace}
\usepackage{mathtools}

\usepackage{multirow,booktabs}
\usepackage{subcaption}

\newcommand{\gaia}{\textit{GAIA}}
\newcommand{\browsecomp}{\textit{BrowseComp}}
\newcommand{\owo}[1]{\textsc{OAgents}}

\definecolor{lightgreen}{RGB}{144, 238, 144} 
\definecolor{lightred}{RGB}{255, 105, 97}   

\newcommand{\roundbox}[2]{%
  \tikz[baseline=(text.base), inner sep=0pt]{
    \node[
      fill=#1,
      text=black,
      minimum height=1.2em,
      rounded corners=3pt,
      inner xsep=4pt,
      inner ysep=1pt
    ] (text) {\small #2};
  }%
}

\newcommand{\uag}[1]{%
  \roundbox{lightgreen}{%
    \scalebox{0.7}{\textcolor{black}{$\uparrow$}}%
    \small\,#1%
  }%
}

\newcommand{\dab}[1]{%
  \roundbox{lightred}{%
    \scalebox{0.7}{\textcolor{black}{$\downarrow$}}%
    \small\,#1%
  }%
}

\newcommand{\fakeparagraph}[1]{\vspace{1mm}\noindent\textbf{#1.}}

\newtcolorbox{promptbox}[2][Prompt]{
colback=black!5!white,
arc=5pt, 
boxrule=0.5pt,
fonttitle=\bfseries,
title=#1, 
before upper={\small}, fontupper=\fontfamily{ptm}\selectfont,
colframe=#2, 
}
\definecolor{ogreen}{RGB}{34, 139, 34}

\usepackage{cleveref}
\theoremstyle{plain}

\theoremstyle{definition}

\theoremstyle{remark}

\usepackage[textsize=tiny]{todonotes}
\usepackage{fontawesome}

\title{OAgents: An Empirical Study of Building \\ Effective Agents}

\affiliation{OPPO AI Agent Team}

\abstract{

Recently, Agentic AI has become an increasingly popular research field. However, we argue that current agent research practices lack standardization and scientific rigor, making it hard to conduct fair comparisons among methods. As a result, it is still unclear how different design choices in agent frameworks affect effectiveness, and measuring their progress remains challenging. In this work, we conduct a systematic empirical study on ~\gaia{} benchmark and \browsecomp{} to examine the impact of popular design choices in key agent components in a fair and rigorous manner. We find that the lack of a standard evaluation protocol makes previous works, even open-sourced ones, non-reproducible, with significant variance between random runs. Therefore, we introduce a more robust evaluation protocol to stabilize comparisons. Our study reveals which components and designs are crucial for effective agents, while others are redundant, despite seeming logical. Based on our findings, we build and open-source \owo{}, a new foundation agent framework that achieves state-of-the-art performance among open-source projects. \owo{} offers a modular design for various agent components, promoting future research in Agentic AI.}

\date{\today}
\correspondence{\\Wangchunshu Zhou at \email{zhouwangchunshu@oppo.com}, Jiaheng Liu at \email{liujiaheng@nju.edu.cn}}
\checkdata[Code]{\url{https://github.com/OPPO-PersonalAI/OAgents}}

\begin{document}
\maketitle

\section{Introduction}
\label{sec:intro}

In recent years, language agents~\cite{autogpt2023autogpt,zhou2023recurrentgpt,wu2023autogen,smolagents,li2023camel,zhou2023agents,zhou2024agents2,xie2023openagents} have received significant attention due to their potential in resolving general, complex tasks that traditionally required human intervention. 
However, despite the surge in the number of research works and open-sourced agent frameworks, current practices in Agentic AI research are far from being rigorous and scientific. 
Specifically, the current landscape of agent research suffers from a lack of standardized designs and implementation details. Critical components such as planning~\cite{yao2023react, shinn2023reflexionlanguageagentsverbal, liu2024RAISE}, memory~\cite{xu2025mem,zhang2024survey,zhou2023recurrentgpt,shi2025taskcraft}, and tool use~\cite{qin2024tool,wang2024executable} vary widely across different papers and frameworks, making it difficult to attribute performance improvements to specific innovations. Compounding this issue, reported results are often hard to reproduce due to inconsistent evaluation settings or undisclosed framework configurations~\cite{owl2025, aworld2025}. This fragmentation undermines the scientific rigor of the field, as findings cannot be reliably compared or built upon. 

Take the widely researched \gaia{} Benchmark~\cite{mialon2023gaia} as an example. Despite the organizers provide a public leaderboard with evaluation code and a number of papers and projects being open-sourced, it is still very hard, if not impossible, for other researchers to reproduce their results because of a number of inconspicuous factors are not standardized, including the implementation details of tools and prompts, as well as details in the evaluation protocol such as how many runs are performed, how errors and failures are handled, and how different results are ensembles or aggregated. These factors often lead to a large impact on the overall performance, sometimes the impact is even larger than some new architecture innovations in new research papers. However, they are generally not mentioned in the technical reports of different agent frameworks and are not even included in their open-sourced codebases. For example, some previous work conducted multiple runs and merged the results, but reported the results as ``pass@1''. Moreover, the engineering design and details in different agent research papers and codebases are so large that it makes it impossible to conduct apples-to-apples comparisons on specific technical designs. This makes it very hard for the research community on agentic AI to properly conduct scientific research instead of digging into tricks on engineering details and evaluation protocols. As a result, despite a lot of agent research papers being released and the numbers on public benchmarks keeping increasing, the best practices on building effective agents are still very obscure.
\begin{figure}[t]
    \centering
    \includegraphics[width=1.0\linewidth]{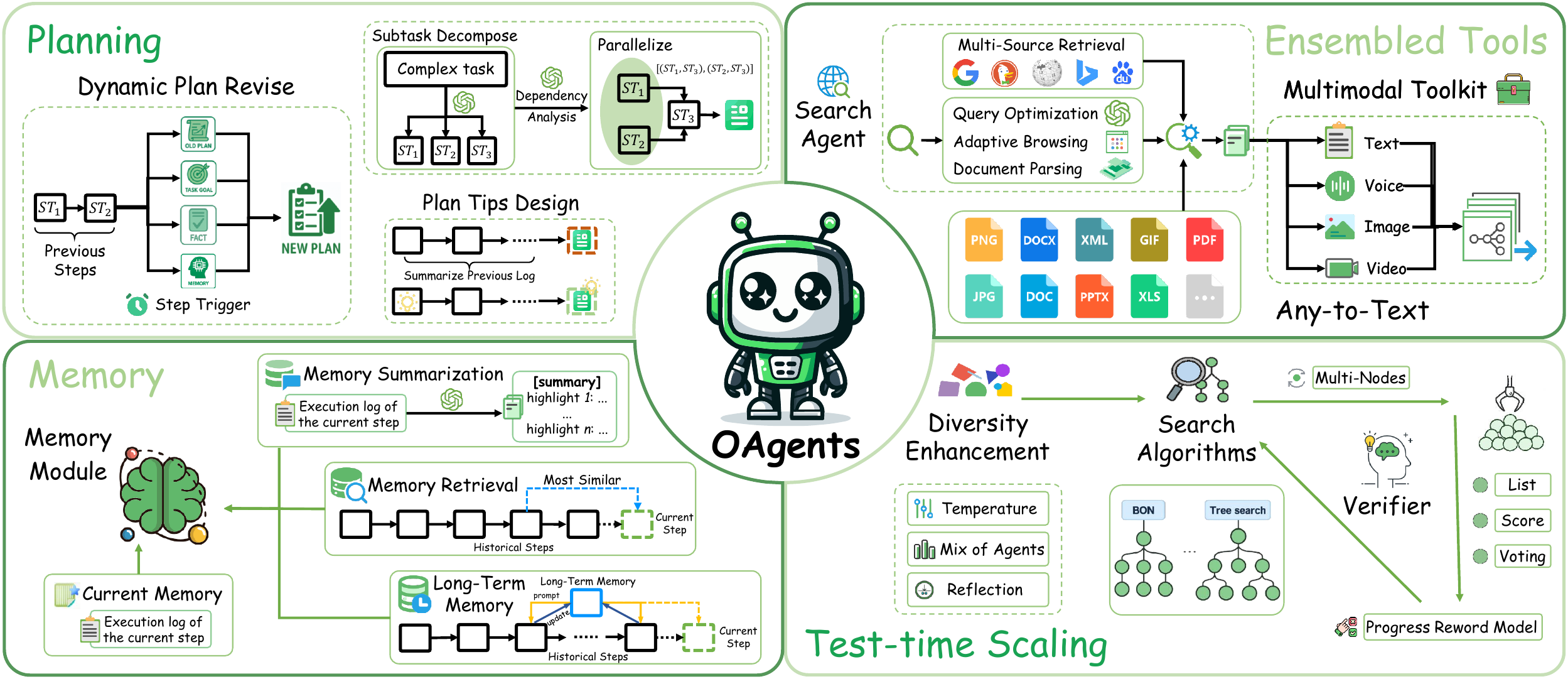}
    \caption{The key components of the \owo{} framework, including planning, memory, tools, and test-time scaling.} 
    \label{fig:framework}
\end{figure}
In this work, to promote truly scientific research on agentic AI and provide researchers with a clear understanding of key factors in building effective agents, we conduct a systematic empirical study on \gaia{} benchmark to sort out the core design choices in current agent research, analyze their impact on performance, and report practical tips for improving experimental stability and reproducibility. Specifically, we: (1) carefully implement and compare different designs on key agent components including planning, tool use, memory, test-time scaling strategies, etc., (2) systematically investigate the impacts of different LLM backbones and their combinations; (3) thoroughly analyze different practices for evaluation and provide a more robust evaluation protocol.

Based on the empirical study, we implement and release \owo{}, a language agent framework that achieves state-of-the-art performance among open-sourced agent frameworks on \gaia{} benchmark and \browsecomp{}. More importantly, \owo{} supports modularized integration of almost all critical designs and features in critical components of language agents, including: (1) different agentic planning mechanisms including static and dynamic workflow designs; (2) a complete tool box including web search with different search sources, browsing tools and web crawlers, parsing tools compatible with more document types. (3) different design of the agentic memory module; (4) test-time scaling strategies for agents including different search algorithms and reflection/self-refine mechanisms. Hopefully, \owo{} will facilitate scientific research on language agents by promoting apple-to-apple comparisons and standardizing evaluation protocols.


In summary, our main contributions are as follows.

\begin{enumerate}
    \item We present a comprehensive agent framework - \owo{}. \owo{} encompass periodically revised plan generation, fine-grained task decomposition \& simultaneous execution, optimization of multi-source web browsing, enhanced document parsing, and adaptive memory mechanisms that collectively enhance performance across various tasks, ranking 1st among open-source agent frameworks on ~\gaia{} benchmark.
    \item We conduct a systematic empirical study and performance analyzes based on the \owo{} framework, offering principles to decompose, analyze and optimize agent designs, uncovering optimal architectural choices and key factors influencing experimental stability.
    \item We introduce practical techniques for reducing experimental variance, including optimization of inference parameters and majority voting strategies, enabling a more reliable and consistent evaluation of agent performance.
\end{enumerate}

\section{Related Work}
\label{sec:related}

\fakeparagraph{Agent Pipeline}
Agents typically undergo iterative planning and execution to accomplish complex tasks, motivating a large amount of research focusing on pipeline design of agents.
~\citet{yao2023react} integrated reasoning and action to simulate the human task execution process, mitigating the issues of hallucination and error propagation, thereby yielding more reliable and interpretable results.
~\citet{autogpt2023autogpt} is a versatile agent framework supporting task decomposition and tool invocation. Moreover, an integrated memory management mechanism endows the Large Language Model (LLM) with long-term and short-term memory capabilities.
~\citet{shinn2023reflexionlanguageagentsverbal} proposed a training-free approach to learn from failure and trial in through linguistic feedback. By recording the observations generated from environment executions, the reflective texts are integrated into the memory, which further guides the planning and execution of subsequent steps.
~\citet{liu2024RAISE} proposed a dual-component memory design, which employs a “Scratchpad” mechanism as short-term memory and maintain a long-term memory repository, significantly enhances the retention and continuity of conversational context.
~\citet{zhou2024languageagenttreesearch} incorporated the Monte Carlo Tree Search (MCTS) method from reinforcement learning to achieve more flexible and adaptive problem-solving, by conducting searches within the combinatorial space of possible reasoning and action steps.

\fakeparagraph{Multi-Agent Systems}
A single agent may encounter performance bottlenecks. Better performance could be achieved by introducing multiple specialized agents. In a multi-agent system, agents typically work as a team or society. ~\citet{guo2024embodied} propose Criticize-Reflect, a hierarchically-organized team within a leader agent and several worker agents. Their study demonstrates that agent teams with a leader are superior than those without a leader. DyLAN~\citep{liu2023dynamic} creates dynamic agent teams that adapt based on past performance, with top contributors reserved for future tasks. The authors demonstrate that re-evaluating and ranking agent contributions would benefit arithmetic and reasoning tasks. 
AgentVerse~\citep{chen2023agentverse} is a multi-agent architecture that enhances reasoning and problem-solving through four key stages: recruitment, collaborative decision making, independent action execution, and evaluation. These stages are repeated iteratively, helping agents reason, collaborate, and act more effectively towards achieving the overall goal.
MetaGPT~\citep{hong2024metagpt} tackles the problem of unproductive chatter in multi-agent architectures by having agents generate structured outputs rather than unstructured chat messages. This approach improves collaboration and focuses the agents' efforts on achieving the team goal.

\fakeparagraph{Agents for GAIA Benchmark.}
GAIA~\cite{mialon2023gaia} presents real-word questions that necessitate fundamental skills including reasoning, handling multiple modalities, web searching, and using various tools. These skills are similar to the requirements of AGI, thus attracting a large number of studies to challenge GAIA. We separate the agent frameworks for GAIA into closed-sourced~\citep{Peng_Langfun_2023,trase2024trase,deepresearch,h2oGPTe2024h2oGPTe,desearch} and open-sourced ones~\cite{owl2025, bahdanau2024tapeagentsholisticframeworkagent, tang2025autoagent, opendeepresearch, smolagents, fourney2024magentic, wu2024copilot}. Smolagents~\cite{smolagents} combines the ReAct~\cite{yao2023react} and Code Act~\cite{wang2024executable} architectures to build a multi-functional agents hierarchy to perform multiple rounds of interactions and actions in code to accomplish complex tasks. Magentic-One~\cite{fourney2024magentic} achieves efficient processing of vision-language tasks
by decoupling perception~\cite{yang2023how2comm,yang2023what2comm},
planning~\cite{song2023llm,tordesillas2021mader},
and execution modules~\cite{qin2024tool,wang2024executable}.
Trase-Agent~\cite{trase2024trase} proposes task reallocation strategies based on real-time feedback,
while TapeAgents~\cite{bahdanau2024tapeagentsholisticframeworkagent}
employs an asynchronous communication framework
to enhance system resilience. AutoAgent~\cite{tang2025autoagent}
enables intelligent task execution and personalized agent creation without coding through the core components
such as natural language-driven multi-agent coordination,
customizable workflows, and self-managing file systems.
Hybrid architecture exploration is exemplified by h2oGPTe-Agent~\cite{h2oGPTe2024h2oGPTe}, which transfers single agent optimization techniques to multi-agent scenarios. Owl~\cite{owl2025} proposes two variants: One is a horizonal architecture named Roleplaying where a user agent asks the questions and assistant agent gives the solutions. The other is a decentralized framework consists of a planner, a coordinator, and specialized workers. Alita~\citep{qiu2025alita} is a concurrent work that achieves excellent results by implementing a self-evolving MCP Box, however the authors did not disclose the algorithm and implementation details of the MCP Box, which is inconsistent with the motivation of our paper.

\section{Building Effective Agents}
\begin{table}[!thb]
\centering
\caption{Performance of various agent frameworks on the \gaia{} benchmark.}
\label{tab:related}
\begin{tabular}{llccccc}
\toprule
\midrule
\textbf{Framework} & \textbf{Model Family} & \textbf{Avg.} & \textbf{Level 1} & \textbf{Level 2} & \textbf{Level 3} \\
\midrule
\multicolumn{6}{l}{\cellcolor[RGB]{221, 204, 212}{\emph{Agentic Model}}} \\

\midrule
Search-o1-32B
& - & 39.8 & 53.8 & 34.6 & 16.7 \\
WebThinker-32B-RL
& - & 48.5 & 56.4 & 50.0 & 16.7 \\
\midrule
\multicolumn{6}{l}{\cellcolor[RGB]{204, 217, 204}{\emph{Closed-source Agent Frameworks}}} \\
\midrule
Langfun Agent    
& Claude-3-7 \etc{} & 71.52 & 83.02 & 68.60 & 57.69 \\
TraseAgent    
& Claude \etc{} & 70.30 & 83.02 & 69.77 & 46.15 \\
Deep Research
& Unknown & 67.36 & 74.29 & 69.06 & 47.60 \\
h2oGPTe   
& Claude-3.5  & 63.64 & 67.92 & 67.44 & 42.31 \\
Desearch         
& GPT-4o & 56.97 & 71.70 & 58.14 & 23.08 \\
\midrule
\multicolumn{6}{l}{\cellcolor[RGB]{198, 230, 204}{\emph{Open‐source Agent Frameworks}} }\\
\midrule
OWL--Workforce              
& Claude-3-7 \etc{} & 69.09 & 84.91  & 67.44 & 42.31 \\
OWL--Roleplaying
& 4o \& o3-mini \etc{} & 58.18 & 81.14 & 54.65 & 23.08 \\
TapeAgents           
& Claude-3-7 \etc{} & 55.76 & 71.70 & 53.49 & 30.77 \\
AutoAgent          
& Claude-3-5 \etc{} & 55.15 & 71.70 & 53.40 & 26.92 \\
Open Deep Research  
& OpenAI o1 & 55.15 & 67.92 & 53.49 & 34.62 \\
Smolagents
& Openai o1 \etc{} & 49.70 & 54.72 & 53.49 & 26.92 \\ 
Magnetic-1
& OpenAI o1 \etc{} & 46.06 & 56.60 & 46.51 & 23.08 \\
FRIDAY       
& GPT-4 turbo  & 34.55 & 45.28 & 34.88 & 11.54 \\
\midrule
\owo{}
& Claude-3-7 \etc{} & 66.67 & 77.36 & 66.28 & 46.15 \\
\owo{}-Pass@3
& Claude-3-7 \etc{} & \textbf{73.93} & \textbf{83.02} & \textbf{74.42} & \textbf{53.85} \\
\midrule
\bottomrule
\end{tabular}
\end{table}
\label{sec:method}
We present a dual-axis analytical paradigm
for architecting cognitive agents in open-world environments,
focusing on two orthogonal evaluation dimensions:
\textit{factual acquisition capacity (FAC)}
and \textit{logical reasoning fidelity (LRF)}.
The FAC axis quantifies an agent's proficiency
in assimilating and updating domain-specific knowledge
from dynamic information streams,
while the LRF axis measures its capability
to maintain rigorous causal relationships
and deduction chains during complex problem-solving.
Through systematic examination of these complementary dimensions,
we establish methodological guidelines for
\textit{1) Enhancing environmental perception through adaptive knowledge integration} and \textit{2) Ensuring decision-making robustness via verifiable inference processes}.
This bifocal approach addresses the fundamental challenges of
balancing empirical learning with formal reasoning
in autonomous artificial systems operating under partial observability.

\noindent \textbf{Factual Acquisition Capacity.}
FAC quantifies an agent's ability to retrieve,
validate, and integrate external knowledge,
fundamentally governed by the \textit{tools} component, which include:

\begin{itemize}
    \item \textbf{Tool Heterogeneity}: Diversity of integrated resources (e.g., search APIs, vision and audio modules) defining accessible knowledge domains.
    \item \textbf{Orchestration Scalability}: Architectural capacity to manage concurrent tool utilization and cross-modal data fusion.
\end{itemize}
Empirical boundaries emerge directly from toolset limitations,
establishing hard constraints on factual knowledge acquisition.

\noindent \textbf{Logical Reasoning Fidelity.}
The LRF framework establishes formal foundations
for stable and coherent decision-making
through synergistic integration of three constitutive elements:
\textit{Plan}, \textit{Memory}, and \textit{Test-Time Scaling}. 
This triadic architecture manifests
distinct operational principles per component:
\begin{itemize}
    \item \textbf{Plan}: Maintains cognitive consistency through temporal synchronization between algorithmic planning strategies and memory-encoded experiential patterns.
    \item \textbf{Memory}: Ensures behavioral coherence through persistent state representations that anchor planning operations across decision episodes.
    \item \textbf{Test-Time Scaling}: Facilitates adaptive resilience by leveraging real-time performance diagnostics to dynamically recalibrate operational parameters.
\end{itemize}

\subsection{Factual Acquisition Capacity (FAC)}
Factual acquisition competence enables agents to
systematically gather, verify, and integrate external knowledge via diverse tools.
This capacity is fundamentally bounded by two critical operational vectors:
multimodal tool interoperability and search tool efficacy,
which jointly define the epistemic frontiers of agent-environment interactions.

We focus on quantifying current capability
ceilings through two investigative lenses:

\begin{itemize}
    \item \textit{Multimodal tool constraints:}
    Characterizing temporal alignment errors and modality fusion bottlenecks
    in cross-domain information synthesis.
    \item \textit{Search tool limitations:}
    Evaluating knowledge coverage gaps imposed by Search API constraints,
    index freshness thresholds, and semantic disambiguation failures in web-scale data retrieval.
\end{itemize}

\subsubsection{Multimodal Toolkit}

To address the limitations in contextual understanding faced
by current agent systems,
a multimodal toolkit is employed that integrates capabilities
for processing text, speech, images, and video.
Unlike traditional frameworks that rely solely on
unimodal conversion to transform
non-textual content into textual descriptions,
this approach enables synchronized and cross-modal semantic parsing:
\begin{equation}
    \text{Response} = \mathcal{A}(x_{\text{text}}, \mathcal{T}_{\text{image}}(I), \mathcal{T}_{\text{video}}(V))
\end{equation}
where \(\mathcal{A}\) is the agent function,
\( x_{\text{text}}\)is the textual input, 
and \(\mathcal{T}_{\text{image}}, \mathcal{T}_{\text{video}}\)
are tool functions that extract features
from images \(I\) and videos \(V\), respectively.
This capability enhances the agent's ability
to acquire and interpret factual information in complex,
real-world scenarios through direct interaction with multimodal inputs.

\subsubsection{Search Agent Framework}
Web search enables LLM-agents
to address real-time information needs and expand epistemic boundaries.
We optimize three subsystems:
\textit{(i) Multi-source retrieval},
\textit{(ii) Query refinement},
and \textit{(iii) Minimalist browsing architecture}
via the Search Agent framework.

\noindent \textbf{Multi-Source Search.}
To mitigate single-source bias,
we integrate commercial APIs (Google, Bing)
and archival systems (Wayback Machine CDX API).
Source selection is state-aware,
driven by query temporal constraints (historical/real-time)
and domain requirements (academic/commercial).
Historical retrieval uses structured ⟨url,date⟩
queries to Internet Archive's temporal index.

\noindent \textbf{Query Optimization Pipeline.}
Closed-loop refinement combines semantic calibration (\textsc{Reflect}) with morphological expansion (\textsc{Expand}):
\begin{equation}
\mbox{\small $
    Q_{\text{opt}} = \textsc{Reflect}(Q_{\text{init}}, M_{\text{task}})
    \rightarrow \textsc{Expand}(Q_{\text{opt}}, L_{\text{term}})
$}
\end{equation}
where \textsc{Reflect}($\cdot$) resolves semantic ambiguities
by calibrating specificity through prompt-based constraints
and logical simplification guided by predefined rewrite rules,
while \textsc{Expand}($\cdot$) generates morphological and semantic variants
via stemming or lemmatization transformations,
as well as domain-specific synonym expansion
(e.g., \emph{COVID-19} $\rightarrow$ \emph{SARS-CoV-2}).

\noindent \textbf{Minimalist Browsing.}
Conventional frameworks suffer from tool overload.
We reduce complexity to three atomic functions:
\texttt{Search (query)}: Find relevant web pages to the query from search engines.
\texttt{Visit (url)}: Navigate to the webpage corresponding to url and \texttt{Read ({url}, {mode})}: Extract contens in a page and present observations.

\subsection{Logical Reasoning Fidelity (LRF)}
\label{sec:reasoning}
In this section,
we investigate three key strategies
to improve logical reasoning in agents:
dynamic plan generation and task decomposition,
memory-augmented knowledge system,
and test-time scaling for exploration optimization.
These approaches address challenges in logical consistency,
environmental adaptability, and efficiency-accuracy trade-offs.

\subsubsection{Dynamic Plan Generation}
\textbf{Strategic Plan Review.}
To enhance agents' complex task management,
planning modules generate high-level plans
\(\mathcal{P} = (s_1, s_2, ..., s_{n})\) 
that decompose tasks into executable steps,
improving reasoning efficiency.
Execution follows the ReAct framework, alternating reasoning
\(r_t\) and actions \(a_t\).
For adaptability in dynamic environments,
plans are revised every $N$ steps using recent observations 
\(\{o_{t-N+1}, ..., o_t \})\):
\(\mathcal{P}^{'} = \text{revise}(\mathcal{P}, \{o_{t-N+1}, ..., o_t \})\) 
This iterative planning-execution loop sustains
goal-directed behavior and strengthens long-term decision-making.

\noindent \textbf{Subtask Decomposition.}
To enhance systematic reasoning in planning modules,
we propose hierarchical task decomposition:
The agent breaks down the main goal
\(\mathcal{G}\) into interdependent subtasks
\(\mathcal{S} = (s_1, s_2, ..., s_{n})\)
and constructs a dependency graph
\(\mathcal{D} = (\mathcal{S}, \varepsilon)\),
where edges \(e_{ij} \in \varepsilon\)
encode precedence constraints.
At each reasoning step \(t\),
dynamic scheduling selects executable subsets
\(\mathcal{S}_{t} \subseteq \mathcal{S}\)
satisfying all dependencies in \(\mathcal{D}\).
Intermediate outputs from completed subtasks
are formalized as structured knowledge representations
\(\mathcal{k}_{i} \in \mathcal{K}\),
which are cross-validated against global constraints \(C(\mathcal{G})\).
A validity function ensures alignment with the overarching goal:
\begin{equation}
    \operatorname{valid}\left(\kappa_{i}\right)=\left\{
    \begin{array}{ll}
    \text{true}, & \text { if } \kappa_{i} \models C(G) \\
    \text{false}, & \text { otherwise }
    \end{array}\right.
\end{equation}

This mechanism enables error detection through consistency checks,
strengthens long-horizon reasoning,
and improves decision-making resilience in complex environments.

\noindent \textbf{Plan Tips.}
To augment planning capabilities,
we propose integrating experiential knowledge
from historical execution trajectories \(\tau\{(s_t, a_t, r_t)\}^{T}_{t}\).
Analysis of past attempts reveals common bottlenecks
and failure patterns,
which are distilled into heuristic guidelines
\(\mathcal{H} = \{h_1, h_2, ...,h_m\}\)
as soft constraints for the planner.
These domain-specific heuristics influence
action selection during planning through an augmented policy:
\begin{equation}
\mbox{\small $
    \small
    \pi_{\theta}\left(a_{t} \mid s_{t}, \mathcal{H}\right)=\operatorname{softmax}\left(Q\left(s_{t}, a_{t}\right)+\beta \cdot f_{\mathcal{H}}\left(s_{t}, a_{t}\right)\right)
    $}
\end{equation}
where \(f_{\mathcal{H}}(\cdot)\) encodes the influence of heuristics
and \(\beta\) controls their weight.
This integration enables preemptive avoidance of known pitfalls,
enhances robustness in plan generation,
and improves adaptability to dynamic environments
by embedding empirical knowledge into decision-making.

\subsubsection{Memory-augmented Knowledge System}

The hierarchical memory module enhances agent cognition through four components: \textit{Current Memory}, \textit{Memory Summarization}, \textit{Vectorized Retrieval}, and \textit{Long-Term Memory}, each addressing distinct aspects of perception and decision-making.

\noindent \textbf{Current Memory.}
Serves as a short-term buffer storing temporally ordered task-specific information \(M^{c}=\{(s_t, a_t)^{t}_{t-\tau}\}\), for real-time processing and on-the-fly decisions.

\noindent \textbf{Memory Summarization.} 
This component transforms raw experience sequences into structured semantic units using topic modeling and sequence-to-sequence generation:
\begin{equation}
    z_{i} = \text{Summarize}(\{(s_t, a_t, r_t)^{t_{i+1}}_{t}\})
\end{equation}
where \(z_{i}\) denotes a memory summarization. By extracting high-salience knowledge, it facilitates efficient downstream processing.

\noindent \textbf{Vectorized Memory Retrieval.}
This component retrieves beneficial historical memories via vector similarity. Specifically, the execution log of each step is embedded into a shared latent space \(\mathcal{E}\): \(\mathcal{E}(x) = \text{Encode}(x)\). Contextually relevant memories are then retrieved based on vector similarity:

\begin{equation}
{{M_{\mathrm{retrieved}}\ =\arg\max\limits_{m\in M}}~\mathrm{sim}(\mathcal{E}(q),\mathcal{E}(m))}
\end{equation}

\noindent \textbf{Long-Term Memory.}
Addresses challenges in lengthy reasoning chains and contextual redundancy during task execution by integrating historical insights. Updates occur through fusion of current memory with existing long-term knowledge, enabling continuous optimization recommendations for task execution.

These components form a structured framework that organizes, stores, and retrieves knowledge at multiple abstraction levels, helping the agent perform effectively in complex environments.


\subsubsection{Test-Time Scaling}
The Test-Time Scaling (TTS)
module enhances agent capabilities through three mechanisms:
diversity enhancement, optimization, and reward modeling.

\noindent \textbf{Diversity Enhancement.}
A mixture-of-agents sampling strategy combines multiple LLM policies 
\(\pi_{\theta_i}\) with weights \(\alpha_i\):
\begin{equation}
    a_t\sim\sum_{i=1}^K\alpha_i\cdot\pi_{\theta_i}\left(\cdot\mid s_t\right)
\end{equation}
This exploits inter-model diversity to generate
broader solution spaces and improve outcome quality.
 
\noindent \textbf{Optimization.}
The TTS module guides agent reasoning through
process-based reward functions 
\(r_t =R(s_t, a_t)\) that assess task progression,
error handling, and efficiency at each step.
Rewards are temporally aggregated as:
\begin{equation}
    R_{\mathrm{total}}=\sum_{t=1}^T\gamma^tr_t
\end{equation}
providing continuous feedback
to refine reasoning trajectories and improve solution accuracy.

\noindent \textbf{Reward Modeling.} 
The TTS module enables real-time reflection for adaptive problem-solving through::
\begin{equation}
    c_t=\mathrm{Reflect}(\{(s_\tau,a_\tau)\}_{\tau=1}^t)
\end{equation}
where \(c_t\) captures corrective insights from past steps, improving error detection and on-the-fly adjustments to enhance overall performance..
\section{Empirical Study} \label{sec:eval}
\subsection{Experimental Setup}
\fakeparagraph{Dataset.}
\gaia{}\cite{mialon2023gaia} presents real-world challenges that demand essential skills like reasoning, handling multi-modal inputs, web browsing, and overall proficiency in tool-calling. True answers are provided for each question, and the correctness of the model response is evaluated with exact match. Due to the instability and randomness of networked experiments, we allow the model to re-answer a question when the answer given by the model is empty or contains ``Unable to determine'' specified in the prompt. However, recalling incorrect answers is illegal. 

\fakeparagraph{Evaluation Protocol.} 
We follow the evaluation protocol of the \gaia{}benchmark~\cite{mialon2023gaia}, which is based on exact match accuracy. The primary metric used is \textit{Pass@N}, which measures the probability that at least one correct solution is found among N independent model attempts. This metric is widely adopted in tasks such as code generation, where the key evaluation criterion is whether the model can produce a valid solution at least once. In our experiments, unless otherwise stated, we report the average \textit{Pass@1} score, reflecting the model's performance in generating a correct answer across all questions within a single evaluation run.

\fakeparagraph{Implementation Details.} 
In both the FAC Evaluations and LRF Evaluations,
the baselines are implemented with
the integrated multi-modal toolkit in \owo{}. For FAC evaluations, we adopt Claude-3-7-Sonnet to activate the code agent and we add plan tips into the system prompts. The web pages are fetched from all 5 types of search engines. While we only set Google as the only search source and disable plan tips for LRF evaluations. 
Unless otherwise specified, all models employed in the agent are based on GPT-4.1 to ensure consistency
in model architecture and capabilities across experiments.

\fakeparagraph{Baseline Methods.}
The selected baselines are categorized into two main groups. Open-source systems include FRIDAY~\cite{wu2024copilot} , Magnetic-1~\cite{fourney2024magentic}, TapeAgent~\cite{bahdanau2024tapeagentsholisticframeworkagent}, AutoAgent~\cite{tang2025autoagent}, Open Deep Research~\cite{opendeepresearch}, and OWL~\cite{owl2025}. Closed-source frameworks comprise Langfun Agent~\cite{Peng_Langfun_2023}, Trase Agent~\cite{trase2024trase}, Deep Research~\cite{deepresearch}, h2oGPTe~\cite{h2oGPTe2024h2oGPTe}, and Desearch~\cite{desearch}. These baselines collectively capture a broad spectrum of current advancements in both open and proprietary multi-agent systems, offering a solid foundation for benchmarking the effectiveness and performance of our proposed \owo{} framework.


    

\subsection{Main Results}
The results in Table~\ref{tab:related} reveal several key insights into the performance landscape across various agent frameworks on the \gaia{} benchmark. Notably, our method (\owo{}-Pass@3) achieves the highest overall average score of 73.93\%, outperforming all other frameworks, including both closed-source and open-source systems. This highlights the robustness and effectiveness of our agent design.

In terms of Level 1 task performance, our method reaches 83.02\%, tying with the best-performing frameworks and establishing a new standard for esay task handling. This superior performance reflects the reliability and consistency of our low-level agents and the underlying System Utilities.When compared with leading closed-source agents like Langfun Agent (71.52\%) and TraseAgent (70.30\%), our method shows a clear edge in both average and Level 2 accuracy. Finally, in the open-source domain, \owo{}-Pass@3 demonstrates a significant margin over the best alternative, OWL-Roleplaying (58.18\%), reaffirming our method's leading position among publicly available systems. Overall, these results validate our approach as a state-of-the-art solution for generalist agent tasks.

We replicate Open Deep Research~\cite{opendeepresearch} and note the results as ``Smolagents'', and the performance of the replication shows a significant degradation. This indicates that the reproducibility of the current agencies framework is poor.

\subsubsection{FAC Evaluations}
\noindent \textbf{Multimodal Toolkit.}
We have refined text extraction tools
with format-specific strategies tailored for various document types
(\textit{pdf}, \textit{xlsx}, and \etc).
For audio inputs, we employ the \textit{whisper-1}
speech-to-text model to generate accurate transcriptions.
For video content, we implement a pipeline combining
keyframe extraction with vision-language models for temporal and contextual analysis.
Importantly, we incorporate a multi-source image understanding module,
which leverages multiple vision language models source
to understand visual features.
Evaluated on the \gaia{} dataset (\Cref{tab:tool}),
our toolkit achieves a cross-modal task accuracy of 74.07\%,
outperforming the baseline system's 48.15\%.
Notably, in audio question-answering subtasks,
temporal reasoning accuracy improves from 0\% to 100\% (3/3).
These results demonstrate that a deeply optimized multimodal
architecture can effectively bridge modality gaps in intelligent agent systems.
\begin{table}[!h]
\centering
\setstretch{1.2}
\caption{%
    Performance ($\%$) of \owo{} before and after integrating multimodal toolkit.
}\label{tab:tool}
\vspace{-9pt}
\begin{tabular}{l|c|ccc}
\toprule
\multirow{2}{*}{Method}
    & \multicolumn{4}{c}{GAIA multimodal tasks} \\
    & Sum & Audio & Image & Tubular \\
    \midrule
Task number & 27 & 3 & 10 & 14 \\
\midrule
\owo{}
    & 48.15 & 0.00 & 40.00 & 64.29\\
\owo{} + \small{Toolkit}
    & 74.07 & 100.00 & 60.00 & 78.57\\
\bottomrule
\end{tabular}
\end{table}

\noindent \textbf{Search Agent.}
Our empirical analysis quantitatively evaluates
how search infrastructure design affects
the performance of \gaia{}.
As shown in~\Cref{tab:crawler}, Jina reader outperforms raw HTML parsing by 9.3\% in Level 2 tasks. Its structured text extraction benefits mid-complexity factual acquisition, highlighting preprocessing’s role in enhancing retrieval quality.
\begin{table}[!h]
\centering
\setstretch{1.1}
\caption{Performance comparison of browser methods on \gaia{} benchmark.
All results are obtained using information retrieved from Google Search.
}
\label{tab:crawler}
\vspace{-9pt}
\begin{tabular}{l|cccc}
\toprule
\multirow{2}{*}{Browser Method}
    & \multicolumn{4}{c}{GAIA} \\
    & Average & Level 1 & Level 2 & Level 3  \\
\midrule
Text web browser    & 44.20 & 54.71 & 43.02 & 26.92 \\
Raw reader          & 49.70 & 64.51 & 46.51 & 30.76 \\
Crawler crawl4ai    & 50.90 & 67.92 & 51.16 & 15.38 \\
Jina reader         & 51.52 & 67.92 & 48.83 & 26.92 \\
\bottomrule
\end{tabular}
\end{table}

From~\Cref{tab:browser}, integrating complementary search engines (DuckDuckGo, Baidu, Bing) consistently improves retrieval accuracy, with the largest gain in Level 3 tasks (+7.69\%). This indicates that diversifying information sources mitigates individual engine limitations, particularly in complex retrieval scenarios.
\begin{table}[!h]
\centering
\setstretch{1.2}
\caption{\owo{} performance of different search source configurations on \gaia{}.
Note that ``single-source'' refers to Google only.
``Multi-source ($k=3$)'' includes Google, Wikipedia, and DuckDuckGo.
``multi-source ($k=5$)'' further adds Bing and Baidu as additional search sources.
}
\vspace{-9pt}
\label{tab:browser}
\begin{tabular}{l|cccc}
\toprule
\multirow{2}{*}{Search Method}
    & \multicolumn{4}{c}{GAIA} \\
    & Average & Level 1 & Level 2 & Level 3  \\
\midrule
Single-source                    & 51.52  & 67.92 & 48.83 & 26.92 \\
Multi-source ($k=3$)             & 52.12  & 67.92 & 50.00 & 26.92 \\
Multi-source ($k=5$)             & 55.15  & 67.92 & 53.49 & 34.61 \\
\bottomrule
\end{tabular}
\end{table}

The proposed query optimization strategy, combining reflection and expansion mechanisms, significantly enhances system performance (\Cref{tab:query_optimization}). It yields a 7.55\% improvement in Level 1 and 2.31\% in Level 2, underscoring the effectiveness of refined query formulation in improving search outcomes.
\begin{table}[!h]
\centering
\setstretch{1.2}
\caption{\owo{} performance comparison of query-optimization configurations on \gaia{}.
}
\label{tab:query_optimization}
\vspace{-9pt}
\begin{tabular}{l|cccc}
\toprule
\multirow{2}{*}{Query Optimization}
    & \multicolumn{4}{c}{GAIA} \\
    & Average & Level 1 & Level 2 & Level 3  \\
\midrule

Raw data                 & 55.15  & 67.92 & 53.49 & 34.61 \\
Reflection-Expansion     & 58.18  & 75.47 & 55.80 & 30.76 \\

\bottomrule
\end{tabular}
\end{table}
Finally, the minimalist system architecture demonstrates competitive performance, supporting the hypothesis that reduced interface complexity can improve robustness without sacrificing functionality.

\noindent \textbf{\owo{}.}
By integrating an optimized search infrastructure
with a multimodal toolkit,
and employing the Jina reader with multi-source $(k=5)$ strategies,
our \owo{} achieves strong improvement
s on the \gaia{} benchmark across diverse base models.
With GPT-4o, \owo{} improves the overall score by 8.09\%,
including a 7.69\% gain in Level 3 tasks.
Gemini-2.5 shows a 9.09\% average improvement,
with Level 3 jumping 19.24\%,
confirming the effectiveness of the multimodal toolkit
and refined search agent.
Notably, Claude-3-7 gains 20.61\%,
the highest observed boost,
demonstrating the framework’s adaptability
to models with varying baseline performance.
The integrated design enhances FAC
through advanced search and multimodal capabilities,
establishing a solid foundation for knowledge-intensive agent systems.
These results confirm that FAC improvements
significantly elevate intelligent agent performance across architectures.
\begin{table}[!h]
\centering
\setstretch{1.1}
\caption{%
   \owo{} performance of various base models on \gaia{}.
}\label{tab:main}
\vspace{-9pt}
\begin{tabular}{l|c|cccc}
\toprule
\multirow{2}{*}{Model} & \multirow{2}{*}{Type}
    & \multicolumn{4}{c}{GAIA Score} \\
    
  & & Average & Level 1 & Level 2 & Level 3 \\
\midrule
\multirow{3}{*}{GPT-4o}
& Baseline & 36.97 & 54.72 & 34.88 & 7.69  \\
& Advance & 45.06 & 62.26 & 45.35 & 15.38  \\
& Gap   & \uag{8.09} & \uag{7.54} & \uag{10.47} & \uag{7.69}  \\
\midrule
\multirow{3}{*}{GPT-4.1}
& Baseline & 44.20 & 54.71 & 43.02 & 26.92  \\
& Advance & 55.15 & 67.92 & 53.49 & 34.62 \\
& Gap   & \uag{10.95} & \uag{13.21} & \uag{10.47} & \uag{7.70}  \\
\midrule
\multirow{3}{*}{OpenAI-o1}
& Baseline & 49.70 & 54.72 & 53.49 & 26.92 \\
& Advance & 53.94 & 67.92 & 52.33 & 30.77 \\
& Gap   & \uag{4.24} & \uag{13.20} & \dab{1.16} & \uag{3.85} \\
\midrule
\multirow{3}{*}{Claude-3-7}
& Baseline & 38.18 & 56.60 & 36.05 & 7.69 \\
& Advance & 58.79 & 64.15 & 61.63 & 38.46 \\
& Gap   & \uag{20.61} & \uag{7.55} & \uag{25.58} & \uag{30.77} \\
\midrule
\multirow{3}{*}{DeepSeek-R1}
& Baseline & 33.90 & 45.28 & 33.72 & 11.54  \\
& Advance & 49.70 & 62.26 & 50.00 & 23.08  \\
& Gap   & \uag{15.80} & \uag{16.98} & \uag{16.28} & \uag{11.54} \\
\midrule
\multirow{3}{*}{Gemini-2.5}
& Baseline & 49.09 & 69.81 & 46.51 & 15.38 \\
& Advance & 58.18 & 73.58 & 55.81 & 34.62  \\
& Gap   & \uag{9.09} & \uag{3.77} & \uag{9.30} & \uag{19.24} \\
\bottomrule
\end{tabular}
\end{table}

\subsubsection{LRF Evaluations}

\textbf{Dynamic Plan Generation.}
The results in Table~\ref{tab:planning} show that our planning and workflow design significantly enhance GPT-4.1’s ability to solve complex tasks.
Strategic plan review (baseline)
improves overall accuracy
by 3.64\% over the static workflow,
confirming that dynamic plan revision
supports better adaptability and long-term reasoning.
Subtask Decomposition achieves a 2.42\%
improvement over baseline,
demonstrating that breaking down tasks into
structured subtasks enhances systematic reasoning,
particularly for tasks of moderate complexity.
The Plan tips are summarized from analysis
of historical error logs and
incorporate heuristic knowledge gained from past failures.
They contribute to a 14.54\% performance improvement,
proving that leveraging prior experience helps prevent errors
and build more robust plans.
This is especially important for high complexity tasks.
Together, these components significantly enhance the system's planning capabilities for complex reasoning.
\begin{table}[!h]
\centering
\setstretch{1.1}
\caption{%
    \owo{} performance evaluation of plan studies on \gaia{}.
    Note that \texttt{Static workflow} refers to a scenario in which
    all tasks follow the same manually designed workflow.
}\label{tab:planning}
\vspace{-9pt}
\begin{tabular}{l|cccc}
\toprule
\multirow{2}{*}{Model combination}
    & \multicolumn{4}{c}{GAIA} \\
    & Average & Level 1 & Level 2 & Level 3  \\
\midrule
\owo{} & 51.52  & 67.92 & 48.83 & 26.92 \\
\ r.p. \texttt{Static workflow}
  & 47.88 &  62.26 & 47.67 & 19.23  \\
\ + \texttt{Subtask}
  & 53.94 & 71.70 & 51.16 & 26.92\\
\ + \texttt{Plan tips}
  & 66.06 & 79.25 & 66.28 & 38.46 \\
\bottomrule
\end{tabular}
\end{table}

\begin{figure}[!thb]
\centering
    \includegraphics[width=.56\linewidth, trim=0.3cm 0.3cm 0 0, clip]{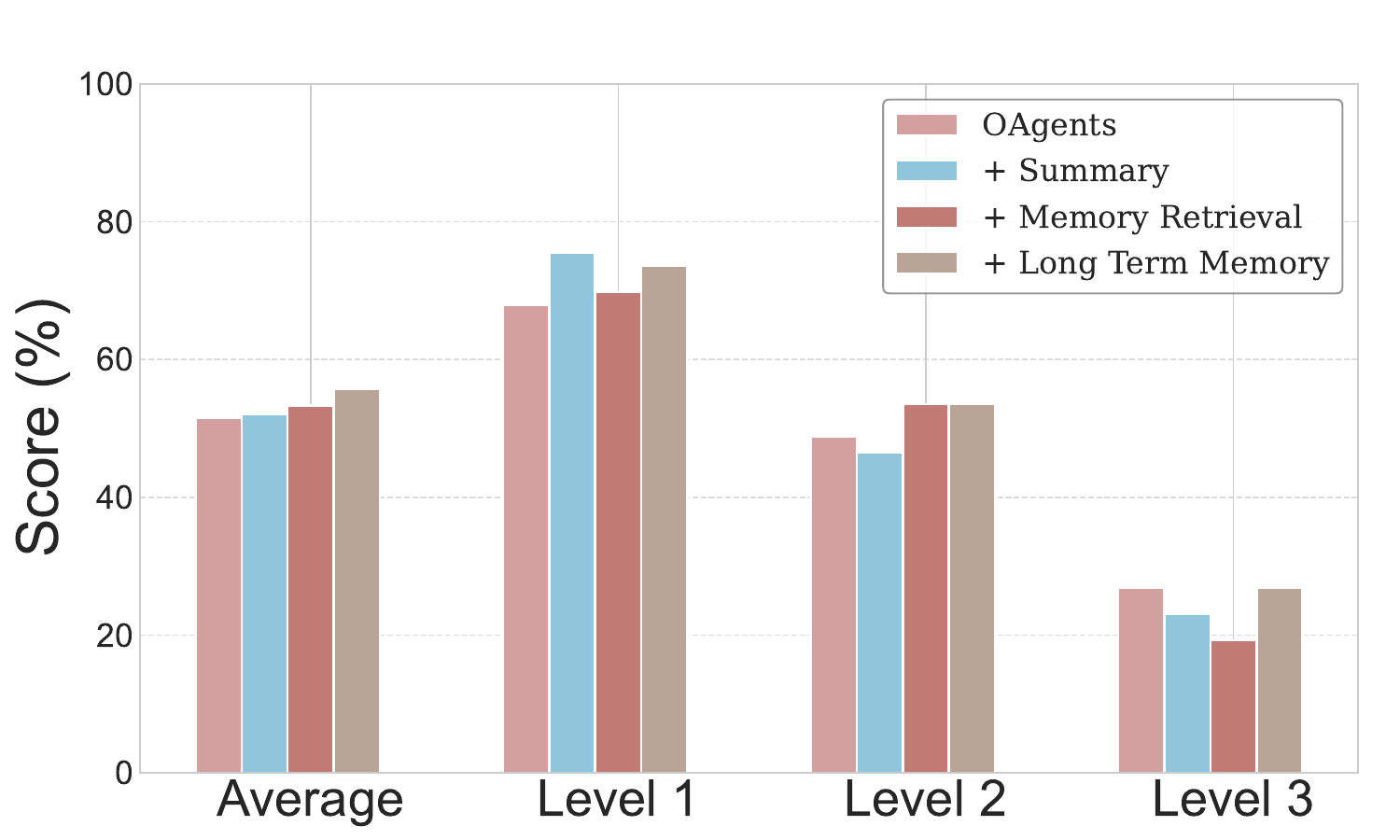}
    \caption{\owo{} performance evaluation of various memory methods on \gaia{}.}
\label{fig:memory}
\end{figure}

\noindent \textbf{Memory.}
The experimental evaluation on the \gaia{} benchmark
highlights the effectiveness of the memory components.
From~\Cref{fig:memory},
adding memory summarization slightly improved
average accuracy from 51.52\% to 52.12\%.
With memory retrieval,
performance increased further to 53.33\%.
The most significant gain came from long-term memory,
raising the average to 55.76\%,
while also achieving the competitive results across all difficulty levels.

Results confirm the benefits of memory components in enhancing agent cognition, future work will explore dynamic component allocation based on task complexity metrics.

\begin{figure}[h]
\centering
    \includegraphics[width=.56\linewidth, trim=0.3cm 0.3cm 0 0, clip]{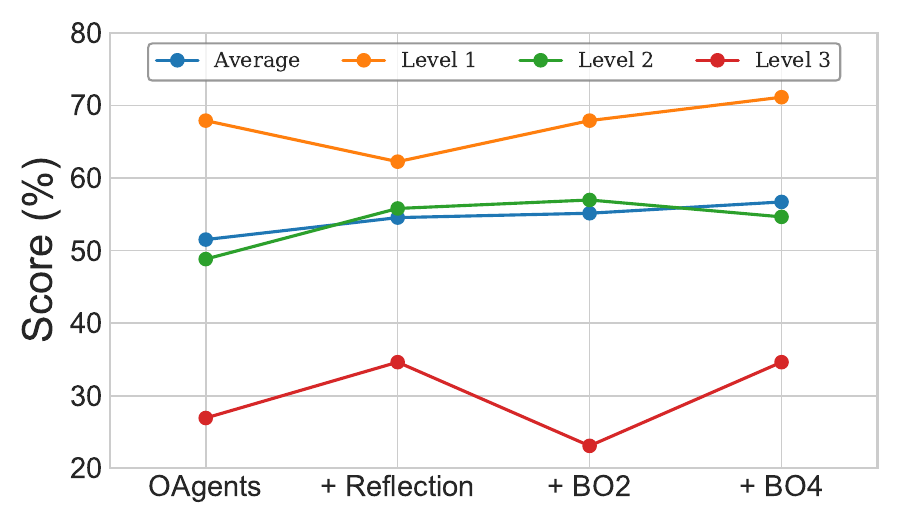}
    \caption{\owo{} performance evaluation of TTS methods on \gaia{}.}
\label{fig:tts}
\vspace{-5mm}
\end{figure}

\noindent \textbf{Test-Time Scaling.}
As shown in Table~\ref{fig:tts},
we conduct an ablation study to examine how test-time scaling (TTS)
strategies influence the performance of \owo{}
across different task complexities.
Reflection leads to a moderate overall improvement (3.03\%),
yet its effects vary across task levels.
While it enhances performance on Level 1 and Level 2 tasks
through iterative reasoning,
it unexpectedly degrades results on Level 3 tasks by 6.62\%,
suggesting potential instability
or error accumulation in complex reasoning chains.
Best-of-N sampling demonstrates more consistent gains,
with performance improving as the sample size increases.
BO2 yields modest improvements (1.82\%),
while BO4 achieves the best overall performance (5.19\%),
particularly benefiting simpler tasks (Level 1: 9.44\%, Level 2: 10.46\%).
This indicates that answer diversification
helps in navigating simpler solution spaces more effectively.

Nonetheless,
neither strategy substantially improves performance on Level 3 tasks,
underscoring the persistent difficulty
in achieving robust multi-step reasoning at scale.
These findings reveal that TTS strategies
exhibit differential effectiveness
depending on task complexity—offering clear benefits
for straightforward tasks
but requiring further innovation to address advanced reasoning challenges.

\begin{table}[!h]
\centering%
\small
\setstretch{1.15}
\caption{%
   \owo{} performance of search agent on BrowserComp-Subset.
}\label{tab:browsercomp}
\vspace{-9pt}
\resizebox{0.46\textwidth}{!}{
\begin{tabular}{l|c}
\toprule
{Model} & {BrowserComp-Subset} \\
\midrule
Claude-3-7 & 4.76\% \\
GPT-4.1 & 7.94\% \\
OpenAI-o1 & 14.29\% \\
\midrule
\owo{} - GPT-4.1 & \textbf{22.22\%} \\
\owo{} - Claude-3-7 & \textbf{22.22\%} \\
\bottomrule
\end{tabular}}
\end{table}
\subsection{Evaluations on \browsecomp}
To validate the search agent's capabilities, we evaluate \owo{} on a more challenging benchmark named \browsecomp{}~\citep{wei2025browsecomp}, where single language models rarely answered correctly or scored. As shown in~\Cref{tab:browsercomp}, \owo{} significantly improved the model's abilities in web browsing.
\section{\gaia{} Benchmark}
\label{sec:ablation}
The \gaia{} benchmark has emerged
as a prominent evaluation framework for
assessing the performance of autonomous agents
in real-world scenarios.
As the leaderboard for this benchmark continues to grow,
it becomes increasingly evident that reported results
often vary in terms of evaluation metrics—particularly
in the use of different \textit{Pass@K} criteria. 
While some methods report \textit{Pass@1},
others adopt more lenient metrics such as \textit{Pass@3}
or even \textit{Pass@5}.
This inconsistency complicates fair comparisons
across different agent frameworks
and limits the transparency of their actual capabilities.
\begin{table}[!h]
\centering
\setstretch{1.2}
\caption{%
    Comparison of performance on the \gaia{} benchmark under different \textit{Pass@K} metrics. 
    Note that "OWL" stands for the open-source role-playing version.
}\label{tab:compare}
\vspace{-9pt}
\begin{tabular}{l|c|c|cccc}
\toprule
\multirow{2}{*}{Method} & \multirow{2}{*}{Model} & \multirow{2}{*}{Metric} & \multicolumn{4}{c}{GAIA}\\
 & &  & Average & Level 1 & Level 2 & Level 3\\
\midrule
\owo{} & Claude-3-7 &\multirow{3}{*}{Pass@1}
& \textbf{66.67} & \textbf{77.36} & \textbf{66.28} & \textbf{46.15} \\
OWL & 4o \& o3-mini & & 53.33
& 71.70 & 50.00 & 26.92   \\
AWorld  & Claude-3-7 & 
& 61.81 & - & - & -  \\
\midrule
\owo{} & Claude-3-7 & \multirow{2}{*}{Pass@3}
& \textbf{73.93} & \textbf{83.02} & \textbf{74.42} & \textbf{53.85}  \\
OWL & 4o \& o3-mini & 
& 58.18 & 81.14 & 54.65 & 23.08  \\
\midrule
AWorld  & Claude-3-7 & {Unknown}
& 77.58 & 88.68 & 77.91 & 53.85\\
\bottomrule
\end{tabular}
\end{table}

To address this issue and ensure alignment with
the leaderboard standards,
we reimplemented the state-of-the-art OWL framework
to obtain its \textit{Pass@1} performance for comparison.
Additionally, we evaluated our proposed open-source framework,
\owo{}, under the \textit{Pass@3} setting,
as summarized in~\Cref{tab:compare}.
Built upon integrated multi-modal toolkit,
multi-source information retrieval,
and test-time scaling (TTS) strategies,
\owo{} demonstrates competitive performance
among existing open-source frameworks under the \textit{Pass@3} metric.
These results highlight the framework’s effectiveness
in handling complex reasoning tasks
and its strong potential for deployment
in real-world applications requiring robust
and scalable reasoning capabilities.
\section{Conclusion}

In this work, we conduct a systematic study on \gaia{} and \browsecomp. We identify key components for effective agents, such as planning, memory, and tool use, and propose a robust evaluation protocol. We release \owo{}, an open-source modular agent framework achieves state-of-the-art performance on \gaia{} (\textbf{73.93}), providing a foundation for future research on agentic agent area.
\newpage
\section{Contributions}
\begin{multicols}{2}
\textbf{Core Contributors}
\begin{itemize}
    \item He Zhu
    \item Tianrui Qin
\end{itemize}
\textbf{Contributors}
\begin{itemize}
    \item King Zhu
    \item Heyuan Huang
    \item Yeyi Guan
    \item Jinxiang Xia
    \item Yi Yao
    \item Hanhao Li
    \item Ningning Wang
    \item Pai Liu
    \item Tianhao Peng
    \item Xin Gui
    \item Xiaowan Li
    \item Yuhui Liu
\end{itemize}

\textbf{Organizers}
\begin{itemize}
\item Yuchen Eleanor Jiang
\item Jun Wang
\item Changwang Zhang
\item Xiangru Tang
\item Ge Zhang
\item Jian Yang
\item Minghao Liu
\item Xitong Gao
\end{itemize}

\textbf{Corresponding Authors}
\begin{itemize}
\item Wangchunshu Zhou
\item Jiaheng Liu
\item[]
\item[]
\item[]
\item[]
\end{itemize}
\end{multicols}







\clearpage
\bibliographystyle{plainnat}
\bibliography{cite}

\newpage
\beginappendix



\section{Details of \owo{}}
\label{app:details_oagent}
\subsection{Search Agent}
Web search constitutes a foundational capability
for LLM-agents to address real-time information needs
and extend their epistemic boundaries.
We focus on optimizing three critical subsystems:
\textit{(i) Multi-source retrieval}, \textit{(ii) Query refinement}, and \textit{(iii) Adaptive browsing} – implemented through the SearchAgent framework.

\noindent \textbf{Multi-Source Search.}
Contemporary search engines exhibit non-overlapping ranking mechanisms and temporal coverage limitations. To mitigate single-source bias, our implementation integrates:
\begin{itemize}
    \item Commercial APIs: Google Custom Search JSON API and Bing Web Search API. 
    \item Archival Systems: Wayback Machine CDX Server API for historical snapshots.
\end{itemize}
In a state-aware routing mechanism, source selection is autonomously driven by:
\begin{itemize}
    \item Query temporal constraints (historical vs. real-time).
    \item Domain-specific coverage requirements (academic vs. commercial).
\end{itemize}

The historical retrieval tool accepts structured inputs as $⟨url, date⟩$ tuples,
querying the Internet Archive's temporal index through:

\lstset{
    language=Python,
    basicstyle=\ttfamily\small,
    keywordstyle=\color{blue},
    commentstyle=\color{green},
    stringstyle=\color{red}, 
    showstringspaces=false,
    numbers=none,
    numberstyle=\tiny\color{gray},
    breaklines=true,
    frame=none
}
\begin{lstlisting}[caption={Example of construct a CDX query to retrieval archive information.}]
def fetch_historical_page(url: str, timestamp: str) -> str:
    cdx_query = f"http://web.archive.org/cdx/search/cdx?url={url}&output=json&from={timestamp}"
\end{lstlisting}

\paragraph{Query Optimization Pipeline.}
The closed-loop query refinement follows:
\begin{equation}
\mbox{\small $
    Q_{\text{opt}} = \textsc{Reflect}(Q_{\text{init}}, M_{\text{task}})
    \rightarrow \textsc{Expand}(Q_{\text{opt}}, L_{\text{term}})
$}
\end{equation}
where \textsc{Reflect}($\cdot$) resolves semantic ambiguities
by calibrating specificity through prompt-based constraints
and logical simplification guided by predefined rewrite rules,
while \textsc{Expand}($\cdot$) generates morphological and semantic variants
via stemming or lemmatization transformations,
as well as domain-specific synonym expansion
(e.g., \emph{COVID-19} $\rightarrow$ \emph{SARS-CoV-2}).

\noindent \textbf{Minimalist browsing architecture.}
Conventional browser emulation frameworks
impose cognitive overhead through excessive tool options.
Our streamlined implementation reduces interaction complexity by:

\begin{itemize}
    \item Eliminate non-essential operations (e.g., click, scroll, find).
    \item Consolidate functionality into three atomic tools: \texttt{Search}, \texttt{Visit}, and \texttt{Read}.
\end{itemize}

\subsection{Strategic Plan Review}
In order to improve an agent’s capability to manage complex tasks,
the incorporation of a planning module is of critical importance. 
Planning module enables the agent to generate
a high-level plan 
\(\mathcal{P} = (s_1, s_2, ..., s_{n})\) 
before execution,
breaking down complex tasks into manageable steps
and improving reasoning efficiency.
Execution typically follows the ReAct framework,
interleaving reasoning and actions:
at each step \(t\), the agent performs either an action \(a_t\)
or a reasoning step \(r_t\).
To ensure adaptability in dynamic environments,
the plan \(\mathcal{P}\) is periodically revised—every
$N$ steps—based on new observations \(o_t\),
updating the sequence of subtasks as
\(\mathcal{P}^{'} = \text{revise}(\mathcal{P}, \{o_{t-N+1}, ..., o_t \})\) 
This iterative approach supports sustained,
goal-directed behavior and enhances the agent’s
long-term reasoning and decision-making capabilities.

\subsection{Subtask Decomposition.}
Given the role of the planning module in managing complex tasks,
we can further consider
a hierarchical task decomposition mechanism
to enhance systematic reasoning.
During planning, the agent decomposes the main goal \(\mathcal{G}\)
into a set of interdependent subtasks \(\mathcal{S} = (s_1, s_2, ..., s_{n})\), and constructs a dependency graph \(\mathcal{D} = (\mathcal{S}, \varepsilon)\), where edges \(e_{ij} \in \varepsilon\)
represent precedence constraints between subtasks.
This structure enables dynamic scheduling
of non-conflicting subtasks at each reasoning step \(t\),
formalized as selecting an executable subset
\(\mathcal{S}_{t} \subseteq \mathcal{S}\)
such that all dependencies in \(\mathcal{D}\) are satisfied.
A key component is the iterative synthesis of intermediate outputs:
results from completed subtasks are formalized as
structured knowledge representations \(\mathcal{k}_{i} \in \mathcal{K}\),
and refined through cross-validation
against global constraints \(C(\mathcal{G})\).
This process ensures alignment with the overarching goal
and supports error detection and correction via consistency checks:
\begin{equation}
    \operatorname{valid}\left(\kappa_{i}\right)=\left\{
    \begin{array}{ll}
    \text{true}, & \text { if } \kappa_{i} \models C(G) \\
    \text{false}, & \text { otherwise }
    \end{array}\right.
\end{equation}
Collectively,
these mechanisms strengthen the planning module’s capacity
for long-horizon reasoning,
enabling more effective
and resilient decision-making in complex environments.

\subsection{Plan Tips.}
Beyond designing diverse planning strategies,
another promising direction lies in enriching
the planning process with additional prior knowledge.
By analyzing the execution trajectories \(\tau\{(s_t, a_t, r_t)\}^{T}_{t}\)
of past attempts, 
w, we can identify common bottlenecks,
failure points,
and suboptimal behaviors encountered by the agent during task realization.
These insights can then be distilled into actionable tips
or heuristic guidelines \(\mathcal{H} = \{h_1, h_2, ...,h_m\}\),
which are subsequently injected into
the planning module as soft constraints or preferences.

Such domain-specific knowledge serves as
supplementary guidance during plan generation,
influencing the selection of actions and subgoals:
\begin{equation}
\mbox{\small $
    \small
    \pi_{\theta}\left(a_{t} \mid s_{t}, \mathcal{H}\right)=\operatorname{softmax}\left(Q\left(s_{t}, a_{t}\right)+\beta \cdot f_{\mathcal{H}}\left(s_{t}, a_{t}\right)\right)
    $}
\end{equation}
where \(f_{\mathcal{H}}(\cdot)\) encodes the influence of heuristics
and \(\beta\) controls their weight.
As a result,
the planner is better equipped to anticipate potential issues,
avoid known pitfalls,
and construct more robust strategies for complex problem-solving.
This integration of experiential knowledge
enhances not only the effectiveness of individual planning steps
but also the overall resilience of the agent
in dynamic and uncertain environments.

\subsection{Memory-augmented Knowledge System}

The hierarchical memory module is designed to enhance the cognitive capabilities of intelligent agents through four complementary components: \textit{Current memory}, \textit{Memory summarization}, \textit{Memory retrieval}, and \textit{Long-term memory}. Each component contributes uniquely to different aspects of perception, reasoning, and decision-making.

\subsubsection{Current Memory.}
As a fundamental default component of the agent, current memory acts as a short-term buffer to capture fine-grained, task-specific information in real time. This buffer maintains recent observations and actions in a temporal sequence  \(M^{c}=\{(s_t, a_t)^{t}_{t-\tau}\}\), enabling the agent to process dynamic environmental inputs with high fidelity and support on-the-fly decision-making.

\subsubsection{Memory Summarization.} 
This component transforms raw experience sequences into structured semantic units using topic modeling and sequence-to-sequence generation:
\begin{equation}
    z_{i} = \text{Summarize}(\{(s_t, a_t, r_t)^{t_{i+1}}_{t}\})
\end{equation}
where \(z_{i}\) denotes a memory summarization. By extracting high-salience knowledge, it facilitates efficient downstream processing.

\subsubsection{Vectorized Memory Retrieval.}
This component retrieves beneficial historical memories via vector similarity. Specifically, the execution log of each step is embedded into a shared latent space \(\mathcal{E}\): \(\mathcal{E}(x) = \text{Encode}(x)\). Contextually relevant memories are then retrieved based on vector similarity:

\begin{equation}
{{M_{\mathrm{retrieved}}\ =\arg\max\limits_{m\in M}}~\mathrm{sim}(\mathcal{E}(q),\mathcal{E}(m))}
\end{equation}

\subsubsection{Long-Term Memory.}
This component is designed to address the challenges of lengthy reasoning chains and redundant contextual information when agents perform tasks by integrating key insights from historical reasoning processes and generating subsequent optimization recommendations. Specifically, the long-term memory component achieves updates by fusing current memory with existing long-term memory, continuously guiding agents in task execution.

\subsection{Test-Time Scaling}
Agent capabilities can be significantly enhanced through the integration of test-time scaling mechanisms, which dynamically refine decision-making, improve adaptability, and promote more robust exploration. Test-Time-Scaling (TTS) module contributes to this enhancement by addressing three core aspects: diversity, optimization, and reward modeling.

\subsubsection{Diversity Enhancement.}
Enhancing the diversity of reasoning paths is crucial for improving agent performance in complex tasks. By leveraging a mixture-of-agents sampling strategy:
\begin{equation}
    a_t\sim\sum_{i=1}^K\alpha_i\cdot\pi_{\theta_i}\left(\cdot\mid s_t\right)
\end{equation}
where \(\alpha_i\) denotes the weight of each agent policy \(\pi_{\theta_i}\),
the TTS module exploits differences in capability profiles across multiple LLMs, generating a broader range of potential solutions and increasing the likelihood of identifying high-quality outcomes.
 
\subsubsection{Optimization.}
To guide agents toward more effective reasoning trajectories, the TTS module introduces process based reward functions \(r_t =R(s_t, a_t)\), which evaluate each step along the generation path. These multi dimensional assessments cover key aspects such as task progression, error handling, and efficiency. The rewards are aggregated over time:
\begin{equation}
    R_{\mathrm{total}}=\sum_{t=1}^T\gamma^tr_t
\end{equation}
providing fine-grained feedback that enables iterative refinement and convergence toward more accurate final responses.

\subsubsection{Reward Modeling.} Real-time reflection and self-correction are essential for adaptive problem-solving. The TTS module incorporates a reflection mechanism that evaluates intermediate steps during exploration:
\begin{equation}
    c_t=\mathrm{Reflect}(\{(s_\tau,a_\tau)\}_{\tau=1}^t)
\end{equation}
where \(c_t\) represents corrective insights fed back into subsequent reasoning stages. This iterative refinement enhances the agent’s ability to detect and rectify errors on-the-fly, leading to improved overall performance.
\newpage
\section{Prompts}
In this section, we post the prompts of essential modules in \owo{} including planning~\S(\ref{apdx:prompt:planning}), search agent~\S(\ref{apdx:prompt:search}), memory~\S(\ref{apdx:prompt:memory}), and test-time scaling~\S(\ref{apdx:prompt:tts}).

\subsection{Planning Prompts}
\label{apdx:prompt:planning}

\resizebox{0.98\textwidth}{!}{
\begin{promptbox}[Plan Tips Prompt]{ogreen}

You are a world expert at making efficient plans to solve any task using a set of carefully crafted tools.

Now for the given task, develop a step-by-step high-level plan taking into account the above inputs and list of facts.\\
This plan should involve individual tasks based on the available tools, that if executed correctly will yield the correct answer.\\
Do not skip steps, do not add any superfluous steps. Only write the high-level plan, DO NOT DETAIL INDIVIDUAL TOOL CALLS.\\
After writing the final step of the plan, write the '<end\_plan>' tag and stop there.\\

Here is your task:\\
\{task\}

You can leverage these tools:\\
\% - for tool in tools.values() \%\\
- \{tool.name\}: \{tool.description\}\\
    \hspace{2em}Takes inputs: \{tool.inputs\}\\
    \hspace{2em}Returns an output of type: \{tool.output\_type\}\\
\%- endfor \%\\

\%- if managed\_agents and managed\_agents.values() | list \%\\
You can also give tasks to team members.\\
Calling a team member works the same as for calling a tool: simply, the only argument you can give in the call is 'task', a long string explaining your task.\\
Given that this team member is a real human, you should be very verbose in your task.\\
Here is a list of the team members that you can call:\\
\%- for agent in managed\_agents.values() \%\\
- \{agent.name\}: \{agent.description\}\\
\%- endfor \%\\
\%- else \%\\
\%- endif \%\\

List of facts that you know:\\
\{answer\_facts\}

Please strictly follow the suggestions below:
\{plan\_tips\}
Now begin! Write your plan below.\\

\end{promptbox}
}

\resizebox{0.98\textwidth}{!}{
\begin{promptbox}[Subtasks Prompt]{ogreen}

You are a world expert at making efficient plans to solve any task using a set of carefully crafted tools.

\textbf{For the given task, you will analyze whether it can be effectively broken down into independent subtasks:}
\begin{itemize}
    \item If the task is more cohesive and cannot be efficiently divided (or would create artificial, highly dependent subtasks), you will treat it as a single task with one comprehensive plan.
    
    \item ELSE if the task can be logically divided into subtasks, you will break it down and provide a separate plan for each subtask.
    \item If the task is more cohesive and cannot be efficiently divided (or would create artificial, highly dependent subtasks), you will treat it as a single task with one comprehensive plan.

    \item ELSE if the task can be logically divided into subtasks, you will break it down and provide a separate plan for each subtask.

    \item Only give the plan, without any explanations or other text.
\end{itemize}

\textbf{IF given task cannot be effectively divided, you will:}
\begin{itemize}
    \item Provide just one plan without the subtask structure.
    \item You will format your response as follows:

1. [First step]

2. [Second step]

...
\end{itemize}
\textbf{ELSE if given task can be broken down into subtasks, you will:}
\begin{itemize}
    \item Identify all subtasks needed to complete the overall task.
    \item Provide a single PARALLEL-LIST that ONLY contains subtasks index joined by comma that can start immediately with NO dependencies on other subtasks.
    \item Provide a complete step-by-step plan for each subtask.
    \item Clearly indicate dependencies between subtasks in the first step of any dependent subtask.
    \item You will format your response as follows:
    
  PARALLEL-LIST [ONLY subtasks index joined by comma that can be started immediately with no dependencies]

  **ST1:[subtask description]
  
  1. [First Step]
  
  2. [Second Step]
  
  ...

  **ST2:[Subtask Description]
  
  1. [First Step - If this subtask depends on another, explicitly state: "Wait for ST[X] to complete" as the first step]
  
  2. [Second Step]
  
  ...

  **STx:[xxx]
  
  ...
\end{itemize}

\textbf{Here is your task:}
\{Task\}

\textbf{List of facts that you know:}
\{Answer\_Facts\}

\textbf{Please strictly follow the suggestions below:}

\{Experience\}\\
Now begin! Write your plan below.\\

\end{promptbox}
}

\subsection{Search Agent Prompts}
\label{apdx:prompt:search}
\resizebox{0.98\textwidth}{!}{
\begin{promptbox}[Query-Reflection Prompt]{ogreen}
You are a highly skilled query evaluation and augmentation agent for a given search query
\begin{itemize}
    \item \textbf{Evaluation Guidline}
    \begin{itemize}
        \item First, identify whether the original query is good enough. Generally, a search query from user may have different problems, for this part, you can refer to Problem Identification.

        \item Second, refer to Available Solution to try to solve the problems that the original query have.

        \item Finally, based on your solution, provide a modified query with your problem analysis and proposed solution.
    \end{itemize}

    \item \textbf{Problem Identification}
    \begin{itemize}
        \item  \texttt{Information Ambiguity}: The entire query is ambiguous, or a part of a long query has semantic ambiguity, which lead to lack of information.
        \item \texttt{Semantic Ambiguity}: The query contains terms with multiple meanings, leading to potential misunderstandings.
        
        \item \texttt{Complex Requirements}: The query involves multiple steps or conditions that need to be addressed separately.
        
        \item \texttt{Overly Specific}: The query is too narrow or detailed, potentially excluding relevant results.

    \end{itemize}

    \item \textbf{Available Solution}
    \begin{itemize}
        \item  \texttt{Information Ambiguity}: Solutions include query expansion or removing the ambiguous part.
        
        \item \texttt{Semantic Ambiguity}: Return the search results for the current query and prompt the model to resolve the ambiguity.
        
        \item \texttt{Complex Requirements}: Break down complex requirements into simpler ones, plan multiple steps, and return the search results for the first step along with the remaining steps to be completed.
        
        \item \texttt{Overly Specific}: Use Less specific query to enhance search quality, remember to retain the core content of the original query.
    \end{itemize}

    \item \textbf{Objective}
    \begin{itemize}
        \item Based on the analysis of the query's issues and the proposed solutions outlined above, you need to optimize and revise the original query
        \item Provide your analysis of the problems, revise suggestions and the final optimized query.
        \item In terms of revising query, please keep your query consice and control the length of new query.
        \item If the original query is already effective with minimal issues, avoid over-revising it.
        \item Attention! You are not allowed to fabricate non-existent information yourself, which is untolerate!
    \end{itemize}

    \item \textbf{Output Format}
    
    You must provide your evaluation in valid JSON format with the following structure:
    \begin{quote}
    \texttt{
    \{
    "Problems": "Consise Analysis of Issues in the Query",
    "Suggestions": "suggestions that help revising the original query"
    "Augmented Query": "your final query"
    \}
    }
    \end{quote}
    \end{itemize}
\end{promptbox}
}

\resizebox{0.98\textwidth}{!}{
\begin{promptbox}[Query-Rollout Prompt]{ogreen}
You are an expert in search strategy and query optimization. Your task is to analyze the user's original query and generate \{roll\_out\} distinct, high-quality search queries.  that:

Please Maintain the core intent of the original query. Read the following guidance.

\textbf{Expand the search scope by incorporating:}
\begin{itemize}
    \item Synonyms or related terms (e.g., "AI" → "artificial intelligence").
    \item Different phrasings (e.g., "causes of climate change" → "climate change drivers").
    \item Domain-specific terminology (e.g., "quantum computing" → "qubit entanglement").
    \item Cross-domain keywords (e.g., "medical AI" → "AI in healthcare diagnostics").
    \item Avoid redundancy while ensuring each query targets a unique angle or sub-topic.

\end{itemize}

\textbf{Query Generate Guidelines:}

\hspace{2em}You are required to Generate \{roll\_out\} queries. Each query should be:

\begin{itemize}
    \item Grammatically correct and concise.
    \item Focused on a specific aspect of the topic.
    \item Appropriately general without exceeding the original's specificity.
    \item Potentially a sub-query of the original.
    \item Different from other queries.
\end{itemize}

\textbf{Input:}

\hspace{2em}Original query: \{query\}

\textbf{Output:}

\hspace{2em}Attention! You are strictly required to provide your final queries within a Python List!

\hspace{2em}The final output \{roll\_out\} queries should in the format as:

\hspace{2em}[query$_1$, query$_2$, ..., query$_N$]
    
\end{promptbox}
}

\subsection{Memory Prompts}
\label{apdx:prompt:memory}
\resizebox{0.98\textwidth}{!}{
\begin{promptbox}[Memory Prompt]{ogreen}
You are an expert in agent memory management, specializing in leveraging the Memory Summarization, the Memory Retrieval, and the Long-term Memory to boost agent reasoning.

\textbf{Memory Summarization:}
\begin{itemize}
    \item Summarize the following text which is the execution content of the agent at the current step: \{memory of current step\}.
    \item Highlight the key points to assist the agent in better reasoning during subsequent steps.
    \item Additionally, you must provide optimization suggestions for the next step.
\end{itemize}

\textbf{Memory Retrieval:}
\begin{itemize}
    \item Summarize the following text, and highlight key points: \{memory of current step\}.
    \item Note that you are only responsible for summarizing, not providing optimization suggestions for the next step.
    \item Your summary of the current step's memory will be vectorized, and then the most relevant historical step memories will be recalled from the vector database by calculating cosine similarity.
\end{itemize}

\textbf{Long-term Memory:}
\begin{itemize}
    \item Here is the agent's execution content from the previous step: \{memory of previous step\}. 
    \item Here is the long-term memory formed by summarizing the agent's historical execution content: \{long term memory\}. 
    \item Please combine the agent's previous execution content and the existing long-term memory, summarize them while highlighting the key points, and form a new long-term memory to help the agent reason better in subsequent steps.
\end{itemize}

\textbf{Input:}
\begin{itemize}
    \item Agent's execution content at current step: \{memory of current step\}.
    
    \item Agent's execution content at previous step: \{memory of previous step\}.
    
    \item Agent's historical execution content: \{long term memory\}.
\end{itemize}

\textbf{Output:}
\begin{itemize}
    \item \textbf{Memory Summarization:} A point-by-point summary of agent's current execution step and optimization suggestions.

    \item \textbf{Memory Retrieval:} The retrieval of the most relevant historical steps.
    
    \item \textbf{Long-term Memory:} An ongoing updated memory for recording the agent's long-term historical steps.
\end{itemize}

\end{promptbox}
}

\subsection{Test-Time Scaling Prompts}
\label{apdx:prompt:tts}
\resizebox{0.98\textwidth}{!}{
\begin{promptbox}[PRM-score Evaluation Prompt]{ogreen}
\textbf{Evaluation Guidelines:}
\begin{itemize}
    \item \textbf{Objective:}
    \begin{itemize}
        \item You will evaluate a candidate ActionStep node, which includes the following fields:
        \begin{itemize}
            \item \texttt{step\_number}: Depth of this step within the TTS search tree.
            \item \texttt{observations}: Observations recorded after executing this action.
            \item \texttt{action\_output}: Direct output resulting from this action.
            \item \texttt{model\_output}: Raw LLM output that led to this action.
            \item \texttt{error}: Any encountered errors (can be None).
            \item \texttt{score}: Previously assigned score (for reference only).
            \item \texttt{previous\_steps}: The history of earlier steps, including TaskStep and PlanningStep, along with the trajectory of ActionSteps leading to the current state.
        \end{itemize}
        \item Your goal is to judge how promising this ActionStep is for advancing toward the user's task, using your independent judgment while considering the continuity and logical flow of the ActionStep sequence, including the historical context.
    \end{itemize}

    \item \textbf{Evaluation Criteria:}
    \begin{itemize}
        \item \textbf{Progress Toward Goal:}
        \begin{itemize}
            \item Assess whether the \texttt{action\_output} clearly and tangibly advances the overall task.
            \item Reward meaningful progress or valuable new information.
            \item Penalize irrelevant actions or weak impact.
        \end{itemize}
        \item \textbf{Error and Stability:}
        \begin{itemize}
            \item Penalize based on the severity of errors:
            \begin{itemize}
                \item Fatal/blocking errors: 0-1 points.
                \item Significant errors: 1-3 points.
                \item Minor or recoverable errors: 3-5 points.
            \end{itemize}
            \item Reduce the score if the \texttt{model\_output} is ambiguous or unstable.
        \end{itemize}
        \item \textbf{TTS Efficiency:}
        \begin{itemize}
            \item Reward actions that contribute efficiently toward reaching the goal.
            \item Penalize redundant or repetitive actions without meaningful progress.
        \end{itemize}
        \item \textbf{Reflection Usage:}
        \begin{itemize}
            \item Reward active utilization of \texttt{reflection} to improve upon past mistakes.
            \item Penalize ignoring reflection insights.
        \end{itemize}
        \item \textbf{Loop Detection:}
        \begin{itemize}
            \item Detect loops or repetitions compared to previous steps.
            \item Identify true loops and penalize based on severity.
        \end{itemize}
        \item \textbf{Contextual Awareness:}
        \begin{itemize}
            \item Infer alignment with previous \texttt{PlanningStep} and \texttt{TaskStep}.
            \item Ensure consistency with the TTS strategy and penalize deviations.
        \end{itemize}
    \end{itemize}

    \item \textbf{Scoring Criteria:}
    \begin{itemize}
        \item \texttt{9-10}: Clearly advances the goal; highly efficient; strong reflection use; no loops.
        \item \texttt{7-8}: Good advancement; minor inefficiencies; clear reflection use; minimal loop risk.
        \item \texttt{5-6}: Moderate progress; limited efficiency; moderate reflection use; mild repetition risks.
        \item \texttt{3-4}: Poor advancement; inefficient; weak reflection use; noticeable loop risks.
        \item \texttt{1-2}: Minimal advancement; repetitive actions; true loops; significant errors.
        \item \texttt{0}: Severe issues: explicit loops, critical errors, or complete irrelevance to the task context.
    \end{itemize}

    \item \textbf{Final Evaluation Output:}
    You must provide your evaluation in valid JSON format with the following structure:
    \begin{quote}
    \texttt{
    \{
  "analysis": "Detailed analysis addressing     progress, TTS efficiency, reflection usage, loop detection, contextual alignment with PlanningStep/TaskStep, error severity, and overall action quality.",
  "score": [integer between 0-10]
    \}
    }
    \end{quote}
    \end{itemize}
\end{promptbox}
}

\resizebox{0.98\textwidth}{!}{
\begin{promptbox}[PRM-list Evaluation Prompt]{ogreen}
\scriptsize 
\textbf{Evaluation Guidelines:}
\begin{itemize}
    \item \textbf{Objective:}
    \begin{itemize}
        \item You will evaluate N candidate trajectories, each representing a series of nodes in a search tree. Each trajectory contains the following:
        \begin{itemize}
            \item \texttt{step\_number}: Depth of the node in the trajectory.
            \item \texttt{observations}: Observations recorded at each step of the trajectory.
            \item \texttt{action\_output}: Direct action output at each step.
            \item \texttt{model\_output}: Raw model output (LLM).
            \item \texttt{error}: Any errors encountered (can be None).
            \item \texttt{score}: Previously calculated score (if available).
            \item \texttt{previous\_steps}: The history of earlier steps, including TaskStep and PlanningStep, with the trajectory of ActionSteps leading to the current state.
        \end{itemize}
        \item Your goal is to evaluate each trajectory holistically, considering how well it progresses toward solving the user's task. Select the trajectory that most effectively achieves this goal.
    \end{itemize}

    \item \textbf{Evaluation Criteria:}
    \begin{itemize}
        \item \textbf{Progress Toward Goal:}
        \begin{itemize}
            \item Assess how well each trajectory advances the task at hand, considering both the individual node's progress and the overall progression of the entire trajectory.
            \item Reward trajectories that demonstrate tangible and meaningful progress toward the goal.
            \item Penalize trajectories with weak actions or minimal/no advancement.
        \end{itemize}
        \item \textbf{Trajectory Efficiency:}
        \begin{itemize}
            \item Evaluate how efficiently each trajectory progresses toward the goal, considering the depth and complexity of the steps.
            \item Favor trajectories that achieve significant progress with fewer steps.
            \item Consider the overall value-to-depth ratio when comparing trajectories of different lengths.
            \item Reward efficient exploration of the search space.
        \end{itemize}
        \item \textbf{Loop Detection:}
        \begin{itemize}
            \item Detect loops or repetitions within each trajectory, especially those related to previous steps.
            \item \textbf{Loop types:}
            \begin{itemize}
                \item \texttt{Real Loops}: Identical nodes (observations, action output, and model output) that do not add value to the trajectory.
                \item \texttt{Benign Repetitions}: Similar strategies with variations yielding additional progress.
            \end{itemize}
            \item Heavily penalize trajectories with real loops.
            \item Slight penalties for benign repetitions if they lead to meaningful improvements.
        \end{itemize}
        \item \textbf{Error and Stability:}
        \begin{itemize}
            \item Evaluate the severity of errors encountered in each trajectory and penalize based on their impact on progression.
            \item \textbf{Error Severity:}
            \begin{itemize}
                \item Fatal/Blocking Errors: Major penalty.
                \item Significant Errors: Moderate penalty.
                \item Minor/Recoverable Issues: Minor penalty.
            \end{itemize}
            \item Penalize unstable or unclear model outputs.
            \item Consider how errors affect the overall trajectory's ability to move toward the goal.
        \end{itemize}
        \item \textbf{Overall Trajectory Quality:}
        \begin{itemize}
            \item Evaluate the coherence and overall quality of the trajectory.
            \item Consider the logical sequence of steps and the exploration-exploitation balance.
            \item Evaluate the final node's closeness to achieving the goal.
            \item Reward trajectories that make consistent progress and demonstrate coherent planning.
        \end{itemize}
    \end{itemize}

    \item \textbf{Final Output Format:}
    Provide your evaluation in the following JSON format. Select the best trajectory and provide a detailed analysis explaining why it is the most promising trajectory.
    \begin{quote}
    \texttt{
    \{
  "index": [integer],  \# Index of the best trajectory
  "analysis": "Detailed analysis addressing progress, efficiency, reflection usage, loop detection, error severity, and overall trajectory quality."
    \}
    }
    \end{quote}
    \end{itemize}

\end{promptbox}
}

\resizebox{0.98\textwidth}{!}{
\begin{promptbox}[Single Node Reflection Prompt]{ogreen}

\textbf{Node Information:}
\begin{itemize}
  \item \texttt{step\_number}: The depth of the node within the BON/beam search tree.
  \item \texttt{observations}: The data or observations recorded during this step.
  \item \texttt{action\_output}: The direct output resulting from an action taken at this step (e.g., API call, tool response).
  \item \texttt{model\_output}: The raw output generated by the model at this step.
  \item \texttt{error}: Any errors encountered during this step (if applicable).
\end{itemize}

\textbf{Goal:}
\begin{itemize}
  \item \textbf{Summarize:}
  \begin{itemize}
    \item Provide a brief overview of what occurred at this node.
    \item Describe the action taken and the results or new information that emerged as a result of this action.
  \end{itemize}

  \item \textbf{Reflect:}
  \begin{itemize}
    \item Assess whether the action taken in this node was successful, partially successful, or unsuccessful.
    \item Identify any errors, issues, or incompleteness relevant to this step.
    \item Compare the node’s outcome with its assigned score, providing an evaluation of whether the score is aligned with the actual result.
  \end{itemize}

  \item \textbf{Confidence:}
  \begin{itemize}
    \item Evaluate your confidence in the action taken at this node (High/Medium/Low).
    \item If confidence is high, explicitly suggest continuing along this exploration path.
    \item If confidence is medium or low, recommend potential improvements or alternatives, while leaving room for exploration to remain open.
  \end{itemize}

  \item \textbf{Suggest:}
  \begin{itemize}
    \item Provide specific and focused suggestions for refining the current step.
    \item These should be based on the evaluation of the current node, with an emphasis on actionable changes that can be made in the next attempt of a similar step.
    \item Focus exclusively on improvements that can be applied within this node. Avoid proposing changes that span multiple steps or introduce larger, long-term strategies.
    \item Base your evaluation strictly on the provided fields—action\_output, observations, error, etc. Do not infer additional context or hypothesize about alternative paths or unknown factors.
    \item Only flag a step as unsuccessful or in need of improvement if there is clear, tangible evidence (e.g., explicit errors, missing or incorrect outputs).
    \item Do not override factual results based on subjective judgment, even if the node's score does not seem to match the outcome.
  \end{itemize}

  \item \textbf{General Guidelines:}
  \begin{itemize}
    \item Your suggestions should be conservative, focusing only on changes where there is a clear issue or opportunity for improvement.
    \item If no significant issues are identified, provide minimal or no suggestions for improvement.
  \end{itemize}
\end{itemize}

\textbf{Output Format:}
\begin{itemize}
  \item \textbf{experience\_summary}: A concise overview of the events at this node and the key outcomes.
  \item \textbf{confidence\_assessment}: High/Medium/Low with a recommendation for future exploration.
  \item \textbf{lessons\_learned}: Key takeaways or specific improvements based on the evaluation of the current node’s action.
  \item \textbf{comments}: Optional minor remarks, clarifications, or additional observations.
\end{itemize}

\end{promptbox}

}
\end{document}